%% file: paper.tex
\documentclass[10pt,twocolumn,letterpaper,dvipsnames]{article}
\usepackage{iccv}
\def\iccvPaperID{192}
\iccvfinalcopy

\usepackage{subfiles}
\usepackage{mathtools}
\usepackage{graphicx}
\usepackage{amsmath,amssymb}
\usepackage{xspace}
\usepackage[pagebackref=true,breaklinks=true,colorlinks,bookmarks=false]{hyperref}
\hypersetup{
  linkcolor=RoyalBlue,
  citecolor=RedViolet,
  urlcolor=RoyalBlue
}
\ificcvfinal\fi
\ificcvfinal\hypersetup{
  pdftitle={Dense Semantic Correspondence where Every Pixel is a Classifier},
  pdfauthor={Hilton Bristow, Jack Valmadre and Simon Lucey},
  pdfsubject={arXiv 2015},
  pdfcreator={LaTeX},
  pdfproducer={IEEE},
  pdfkeywords={SIFT, Flow, LDA, dense correspondence, arXiv, 2015}
}
\fi

\input{notation}

\title{Dense Semantic Correspondence where Every Pixel is a Classifier\vspace{-3mm}}
\author{
  \href{mailto:hilton.bristow@gmail.com}{Hilton Bristow},\textsuperscript{1}
  \href{mailto:jack.valmadre@gmail.com}{Jack Valmadre}\textsuperscript{1} and
  \href{mailto:slucey@cs.cmu.edu}{Simon Lucey}\textsuperscript{2} \\
  \textsuperscript{1}Queensland University of Technology, Australia \\
  \textsuperscript{2}Carnegie Mellon University, USA
}

\begin{document}
\maketitle


\subfile{1-abstract}

\subfile{2-introduction}

\subfile{3-method}

\subfile{4-evaluation}

\subfile{5-conclusion}

\footnotesize
\bibliographystyle{ieee}
\bibliography{citations}
\end{document}

%% file: notation.tex
\newcommand{\fig}[1]{Figure \ref{fig:#1}}

\newcommand{\eqn}[1]{Equation (\ref{eqn:#1})}

\newcommand{\f}{f}
\newcommand{\g}{g}
\newcommand{\h}{h}

\newcommand{\I}{\mathcal{I}}

\newcommand{\R}{\mathcal{R}}
\newcommand{\W}{\mathcal{W}}

\newcommand{\w}{\mathbf{w}}
\newcommand{\x}{\boldsymbol{x}}
\newcommand{\y}{\mathbf{y}}

\renewcommand{\S}{\mathbf{S}}
\renewcommand{\b}{\mathbf{b}}
\renewcommand{\u}{\bar{\x}}

\newcommand{\abs}{\textrm{abs}}

\newcommand{\conv}{\ast}

\newcommand{\e}{\mathrm{e}}
\newcommand{\eye}{\mathbf{I}}

\renewcommand{\neg}{\textrm{neg}}

\newcommand{\pos}{\textrm{pos}}

\newcommand{\fmin}{\textrm{min}}

\makeatletter
\DeclareRobustCommand\onedot{\futurelet\@let@token\@onedot}
\def\@onedot{\ifx\@let@token.\else.\null\fi\xspace}

\renewcommand{\eg}{\emph{e.g}\onedot}
\renewcommand{\ie}{\emph{i.e}\onedot}

\renewcommand{\etal}{\emph{et al}\onedot}

\makeatother

%% file: 1-abstract.tex
\abstract{
Determining dense semantic correspondences across objects and scenes is a difficult problem that underpins many higher-level computer vision algorithms. Unlike canonical dense correspondence problems which consider images that are spatially or temporally adjacent, semantic correspondence is characterized by images that share similar high-level structures whose exact appearance and geometry may differ.

Motivated by object recognition literature and recent work on rapidly estimating linear classifiers, we treat semantic correspondence as a constrained detection problem, where an exemplar LDA classifier is learned for each pixel. LDA classifiers have two distinct benefits:
(i) they exhibit higher average precision than similarity metrics typically used in correspondence problems, and
(ii) unlike exemplar SVM, can output globally interpretable posterior probabilities without calibration, whilst also being significantly faster to train.

We pose the correspondence problem as a graphical model, where the unary potentials are computed via convolution with the set of exemplar classifiers, and the joint potentials enforce smoothly varying correspondence assignment.

}

%% file: 2-introduction.tex
\begin{figure}[t]
\includegraphics[width=\columnwidth]{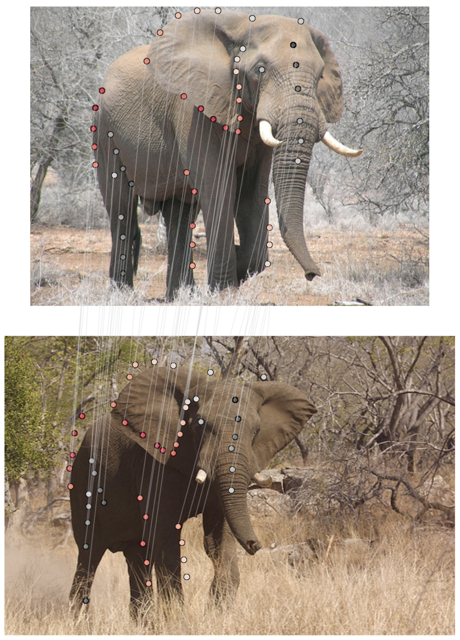}
\caption{Dense semantic correspondence estimates how points are related between images that stem from the same visual class. Here, we wish to predict where each pixel on the first elephant corresponds to on the second, whilst being robust to appearance, pose and background variation. The points labelled are representative of the dense correspondence field estimated by our method.
\label{fig:main}
}
\end{figure}

\section{Introduction}
\label{sec:introduction}
Unlike canonical dense correspondence problems which consider images that are spatially (stereo) or temporally (optical flow) adjacent, semantic correspondence is characterized by images that stem from the same visual class (\eg elephants, lammergeiers, car-lined streets) whilst exhibiting individual appearance and geometric properties.

For example, given two images of elephants (see \fig{main}), we would like to predict where each pixel on the first elephant corresponds to on the second. This is particularly challenging because the space of elephants exhibits significant intra-class appearance and geometric variation. A related problem is that of pose estimation~\cite{ramakrishna_ECCV_2014,yang_CVPR_2011}, which considers a smaller fixed set of landmarks stemming from a labelled dataset of a known object class. From this dataset, one can learn (i) the geometric dependency between landmarks, and (ii) local detectors that discriminate the appearance of each landmark from the background. When presented with a new image, one can then estimate the landmark locations by solving a graphical inference problem.

Liu~\etal's seminal work of SIFT Flow~\cite{liu_PAMI_2011} established that a similar strategy could be applied for estimating dense semantic correspondence between two images stemming from the same semantic class. There are three complicating factors however: (i) learning geometric dependencies between landmarks is impossible from only a single example, (ii) learning local detectors is problematic due to the lack of positive training samples, and (iii) computational complexity is a major concern as we are treating each pixel coordinate within the image as a landmark. Liu~\etal proposed to circumvent these problems by assuming the dense geometric dependencies in an image can be adequately governed by a variational regularizer, and that accurate local detections between semantically similar images can be attained through the $L_1$ distance between SIFT descriptors. Since there is no learning required, this can be performed in a computationally tractable manner.

In this paper, we explore the possibility of actually learning a discriminative detector at every pixel coordinate in an image. Motivated by object detection literature, we learn a linear classifier per pixel in the reference image and apply it in a sliding-window manner to the target image to produce a match likelihood estimate. Learning a multitude of linear detectors such as exemplar support vector machines (SVMs) has typically had two issues: (i) each detector must parse the negative set, often with hard-negative mining techniques, leading to long training times, which makes training a classifier for every pixel in an image intractable, and (ii) since the scale of the outputs depends on the margin, the output confidences of two different SVMs are not directly comparable.

We leverage recent work on learning detectors quickly with linear discriminant analysis (LDA), by collecting negative statistics across a large number of images in a pre-training phase. Learning a new exemplar detector then involves a single matrix-vector multiplication. Since LDA uses a generative model of the class distributions, the posterior probabilities provide a quantity that is comparable between detectors. This allows us to estimate both the likelihood of matches for each pixel individually, and also a global belief of match quality.

\section{Prior Art}
\label{sec:prior-art}

Canonical correspondence problems such as stereo and optical flow typically rely on simple (dis-)similarity metrics to describe the likelihood of two pixels matching. In the original work of Horn and Schunck~\cite{horn_AI_1981}, this was Euclidean distance on raw pixel intensities, which manifested a brightness constancy assumption.

Since then, significant literature has focused on determining robust metrics under increasingly adverse conditions - from non-rigid deformations and occlusions, to non-global intensity, constrast and colorimetric changes~\cite{brox_ECCV_2004,mileva_PR_2007,seitz_ICCV_2009,sun_ECCV_2008}. Importantly, however, all of these works assume the images being observed stem from the same underlying scene.

SIFT Flow first introduced the idea of semantic correspondence \emph{across} scenes~\cite{liu_PAMI_2011}. While the method uses a simple $L_1$ metric, the images are represented in dense SIFT space typically associated with sparse keypoint matching.\footnote{Feature representation and similarity metric are intrinsically related, since $f(\phi(\x_1), \phi(\x_2)) = g(\x_1, \x_2)$.} This sacrifices some localization accuracy for improved geometric invariance.
We maintain, however, that similarity metrics are insufficient for estimating the likelihood of pixels matching between different scenes. Instead, we advocate the use of classifiers, as per deformable face fitting and pose estimation literature, except where a classifier is trained~\emph{per pixel}.

We leverage recent work on rapid estimation of LDA classifiers~\cite{hariharan_ECCV_2012, valmadre_ACCV_2014} to achieve this goal, though fast correlation filter estimation~\cite{henriques_PAMI_2014} is potentially equally applicable. The method we present is largely agnostic to the objective used to learn the linear detectors (\eg SVM, LDA, correlation filters), however LDA classifiers are attractive in producing globally interpretable outputs across pixels, and requiring only a single matrix-vector multiplication to train, which is critical to learning $> 10,000$ classifiers per image.

A number of dense correspondence methods have made use of discriminative pre-training~\cite{li_ECCV_2008,roth_ICCV_2005,sun_ECCV_2008}, with the recent work of~\cite{ladicky_arxiv_2015} being particularly relevant to our discussion. In this work, a classifier of the form $f(\Phi(\x_1) - \Phi(\x_2))$ is trained to predict a (binary) likelihood of two pixels matching. Intuitively, the classifier learns the modes and scale of variation in the underlying feature space $\Phi$ that are important and those that are distractors. Training is fully supervised from groundtruth optical flow data.

Like SIFT Flow,~\cite{ladicky_arxiv_2015} formulate the correspondence objective as a graphical model (\cite{kolmogorov_PAMI_2006,krahenbuhl_NIPS_2011} respectively). This has the distinct advantage over variational methods of permitting very large displacements and arbitrarily complex data terms, at the expense of requiring simple regularizers to keep inference tractable. More recently, a number of variational methods have used sparse descriptor matching to anchor larger displacements~\cite{brox_CVPR_2009,weinzaepfel_ICCV_2013}. While both methods use robust SIFT descriptors for keypoint matching, in a semantic correspondence setting the best match is infrequently the true correspondence, leading to poor initialization of the densification stage.

%% file: 3-method.tex
\section{Dense Semantic Correspondence}
\label{sec:methodology}
Given two images, $\I_A \in \R^{MN}$ and $\I_B \in \R^{PQ}$, and a discrete set of points $\x$, dense semantic correspondence involves minimizing the inverse fitting problem,
\begin{align}
\x^{*} = \arg\min \sum_{i=1}^{MN} \; \f_i(\x_i) + \lambda \g(\x) \label{eqn:correspondence}
\end{align}
where $\f$ is the unary function that evaluates the likelihood of a particular assignment for each $\x_i$ based on the image content, and $\g$ is a regularizer which enforces constraints on the joint configuration of the points. In semantic correspondence, the unary function must be a good indicator of semantic similarity, and so must be robust to significant intra-class variation. In the framework we adopt, there are no constraints on its complexity or properties.

SIFT Flow~\cite{liu_PAMI_2011} adopts a unary of the form,
\begin{align}
\f_i(\x_i) = \h(i, \x_i) = || \Phi_A(i) - \Phi_B(\x_i) ||_1
\end{align}
where $\Phi_A(\x_i) = \Phi(\x_i; \I_A)$ is a feature representation of the image $\I_A$ evaluated at the point $\x_i$.\footnote{For our LDA classifiers, we extract features from a window of pixels around $\x_i$, but this detail can be subsumed into the feature transform $\Phi$.} 

In~\cite{ladicky_arxiv_2015}, the $L_1$ norm on the difference between features is replaced with a more general learned representation,
\begin{align}
\h(i, \x_i) = H(\Phi_A(i) - \Phi_B(\x_i))
\end{align}
In both formulations, however, the unary function is a stationary kernel. This implies a feature space capable of producing similar outputs for semantically similar inputs. Finding such a feature embedding is a difficult task in general, and as a result significant object detection literature has focussed on learning classifiers to distinguish classes instead.

The use of classifiers has two distinct advantages over stationary kernels for describing match likelihood. First, linear classifiers define half-spaces in which samples are either classified as positive or negative. Thus two points with dissimilar appearances can still be afforded a high match likelihood. Second, the importance of different dimensions in the feature space can be learned from data. 

In this paper, we advocate a unary function of the form,
\begin{align}
\f_i(\x_i) = \h(i, \x_i) = \w_A(i)^T \Phi_B(\x_i)
\end{align}
where $\w_A(i)$ is a linear classifier trained to predict correspondences to pixel $i$ in $\I_A$, with ideal response,
\begin{align}
\w_A(i)^T \Phi_B(\x_i) = \left\{
\begin{array}{ll}
\phantom{-}1 & \x_i = \x_i^* \\
-1 & \textrm{otherwise}
\end{array}
\right.
\end{align}

This is traditional binary classification, where the positive class contains the reference pixel, and its true correspondence in the target image, and the negative class contains all other pixels. Since the correspondence in the target image is not known \emph{a priori} however, we rely on the classifier $\w_A(i)$ to generalize from a single training example: $\Phi_A(i)$. This is known as exemplar-based classification~\cite{malisiewicz_ICCV_2011}.

The challenge is how to rapidly estimate thousands of exemplar classifiers per image in reasonable time. The remainder of this section focuses on addressing that challenge, and a number of interesting properties that arise from our approach.

\subsection{Learning Detectors Rapidly using Structured Covariance Matrices}
Linear classifiers have a rich history in computer vision, not least because of their interpretation and efficient implementation as a convolution operation. Support vector machines have proven particularly popular, due to their elegant theoretical interpretation, and impressive real-world performance, especially on object and part detection tasks.
A challenge for any object detection problem is how to treat the potentially infinite negative set (comprising all incorrect correspondences in our case). Object detection methods using support vector machines employ hard negative mining strategies to search the negative set for difficult examples, which can be represented parametrically in terms of the decision hyperplane. This feature is also their limitation for rapid estimation of many classifiers, since each classifier must reparse the negative set looking for hard examples -- knowing one classifier does not help in estimating another.\footnote{This is not strictly true. Warm starting an SVM from a previous solution, especially in exemplar SVMs where only a single positive example changes, can induce a significant empirical speedup, however is unlikely to change the $O()$ complexity of the algorithm.}

Linear Discriminant Analysis (LDA), on the other hand, summarizes the negative set into its mean and covariance. The parameters $\w$ of the decision hyperplane $\w^T \x = c$ are learned by solving the system of equations,
\begin{align}
\S\w = \b \label{eqn:lda}
\end{align}
where $\S$ is the joint covariance of both classes and $\b = \u_\pos - \u_\neg$ is the difference between class means.
\cite{hariharan_ECCV_2012} made two key observations about LDA: (i) if the number of positive examples is small compared to the number of negative examples, the joint covariance $\S$ can be approximated by the covariance of the negative distribution alone, and reused for all positive classes, and (ii) gathering and storing the covariance can be performed efficiently if the negative class is shift invariant (\ie a translated negative example is still a negative example).

This second fact implies stationarity of the negative distribution, where the covariance of two pixels is defined entirely by their relative displacement. Importantly, both \cite{hariharan_ECCV_2012} and \cite{henriques_ICCV_2013} showed that the performance of linear detectors learned by exploiting the stationarity of the negative set is comparable to SVM training with hard negative mining.

The covariance $\S$ can be constructed from a relative displacement tensor, according to,
\begin{align}
\S_{(u,v,p),(i,j,q)} = g[i-u,j-v,p,q] \label{eqn:statistics}
\end{align}
where $i,j,u,v$ index spatial co-ordinates, and $p,q$ index channels. We call the maximum displacement observed $\abs(i-u)$, $\abs(j-v)$ the bandwidth of the tensor. Also note that stationarity only exists spatially -- cross-channel correlations are stored explicitly. The storage of $g$ thus scales quadratically in both bandwidth and channels, though since the detectors we consider are typically small-support, we can entertain feature representations with large numbers of channels (\ie SIFT).

In order to compute $\g$, we gather statistics across a random subset of $50,000$ images from \mbox{ImageNet}. We precompute the covariance matrix of the chosen detector size (typically $5 \times 5$) and factor it with either a Cholesky decomposition, or its explicit \mbox{inverse}, making sure the covariance is positive-definite by adding the minimum of zero and the minimum eigenvalue to the diagonal, \ie $(\S + \fmin(0, \lambda_\fmin) \cdot \eye)^{-1}$.

For each pixel in the reference image, we compute,
\begin{align}
\w_A(i) = \S^{-1} ( \u_\pos - \u_\neg )
\end{align}
which involves a single vector substraction and matrix-vector multiplication, where,
\begin{align}
\u_\pos = \Phi_A (\x_i)
\end{align}

The likelihood estimate for the $i$-th reference point across the target image can be performed via convolution over the discretize pixel grid,
\begin{align}
\f_i(\x) = \w_A(i) \conv \Phi_B(\x) \label{eqn:conv}
\end{align}
Since storing the full unary is quadratic in the number of image pixels (quartic in the dimension), we perform coarse-to-fine or windowed search as per SIFT Flow~\cite{liu_PAMI_2011}.

\subsection{Posterior Probability Estimation}
Linear Discriminant Analysis (LDA) has the attractive property of generatively modelling classes as Gaussian distributions with equal (co-)variance. This permits direct computation of posterior probabilities via application of Bayes' Rule:
\begin{align}
p(C_\pos | \x) = \frac{ p(\x | C_\pos) \; p(C_\pos) }{ \sum\limits_{\mathclap{n \in \{\pos,\neg\}}} p(\x | C_n) \; p(C_n) } \label{eqn:posterior}
\end{align}
where,
\begin{align}
p(\x | C_n) = \frac{1}{(2\pi)|\S|^{\frac{1}{2}}}\e^{-\frac{1}{2} (\x - \u_n)^T \S^{-1} (\x - \u_n)}
\end{align}
With some manipulation, the posterior of \eqn{posterior} can be expressed as,

\begin{alignat}{4}
&p(C_\pos | \x) &&= \frac{1}{1 + \e^{-\y}} \label{eqn:logistic} \\
&\y &&= \x^T \S^{-1}(\u_\pos - \u_\neg) \label{eqn:lda2} \\
&&&+ \tfrac{1}{2}\u_\pos^T \S^{-1}\u_\pos - \tfrac{1}{2}\u_\neg^T \S^{-1}\u_\neg \label{eqn:bias} \\
&&& + \ln \left( \frac{p(C_\pos)}{p(C_\neg)} \right) \label{eqn:prior}
\end{alignat}
\eqn{logistic} takes the form of a logistic function, which maps the domain $(-\infty \dots \infty)$ to the range $(0 \dots 1)$.

The logistic function is typically used to convert SVM outputs to probabilistic estimates, however a ``calibration'' phase is required to learn the bias and variance of each SVM in the ensemble so their outputs are comparable. With LDA, these parameters are derived directly from the underlying distributions.

\eqn{lda2} is the canonical response to the LDA classifier, \eqn{bias} represents the bias of the distributions, and \eqn{prior} is the ratio of prior probabilities of the classes. This must be determined by cross-validation (once, not for each classifier), based on the desired sensitivity to true versus false positives.

\begin{figure*}[t]
\includegraphics[width=\textwidth,trim=0 56pt 0 56pt,clip]{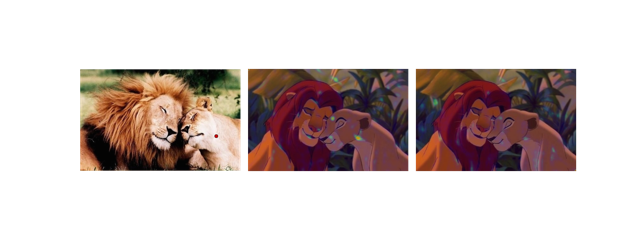}
\includegraphics[width=\textwidth,trim=0 56pt 0 56pt,clip]{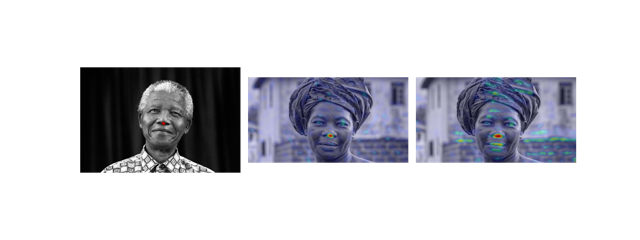}
\includegraphics[width=\textwidth,trim=0 56pt 0 56pt,clip]{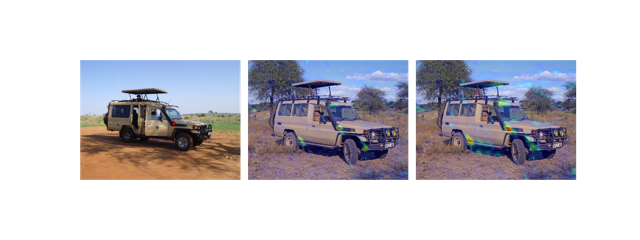}
\includegraphics[width=\textwidth,trim=0 56pt 0 56pt,clip]{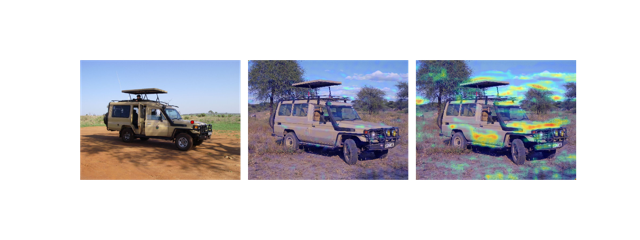}
\includegraphics[width=\textwidth,trim=0 56pt 0 56pt,clip]{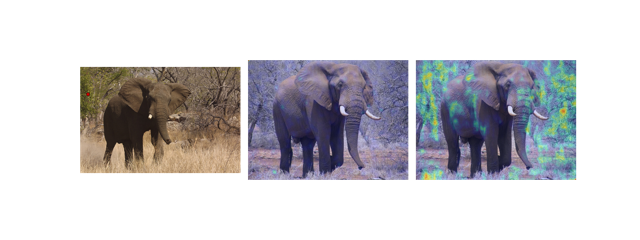}
\caption{From left to right: (a) reference image with reference point labelled in red, and posterior estimates for (b) LDA and (c) $L_1$ norm. We present a range of points, from distinctive to indistinctive or background. LDA and $L_1$ norm have similar likelihood quality for distinctive points, but LDA consistently offers better rejection of incorrect matches and background content.
\label{fig:pde}}
\end{figure*}

By completing the squares in \eqn{bias}, we yield the final expression for computing the posterior probability,
\begin{alignat}{4}
&\y &&= (\x - \tfrac{1}{2}(\u_\pos + \u_\neg))^T \S^{-1} (\u_\pos - \u_\neg) + \mu \nonumber\\
&&&   = (\x - \tfrac{1}{2}(\u_\pos + \u_\neg))^T \w_i + \mu \label{eqn:lda_probability}
\end{alignat}

The implication of \eqn{lda_probability} is that it is no more expensive to compute probability estimates than to just evaluate the classifier -- the computation is still dominated by the single matrix-vector product required to learn the classifier.

\fig{pde} illustrates a representative set of likelihood estimates output by our method and SIFT Flow respectively. LDA typically has tighter responses around the true correspondence, and better suppression of false positives, especially on background content that has no clear correspondence.

%% file: 4-evaluation.tex
\section{Evaluation}
\label{sec:dataset}
\begin{figure*}
\includegraphics[width=0.2\textwidth]{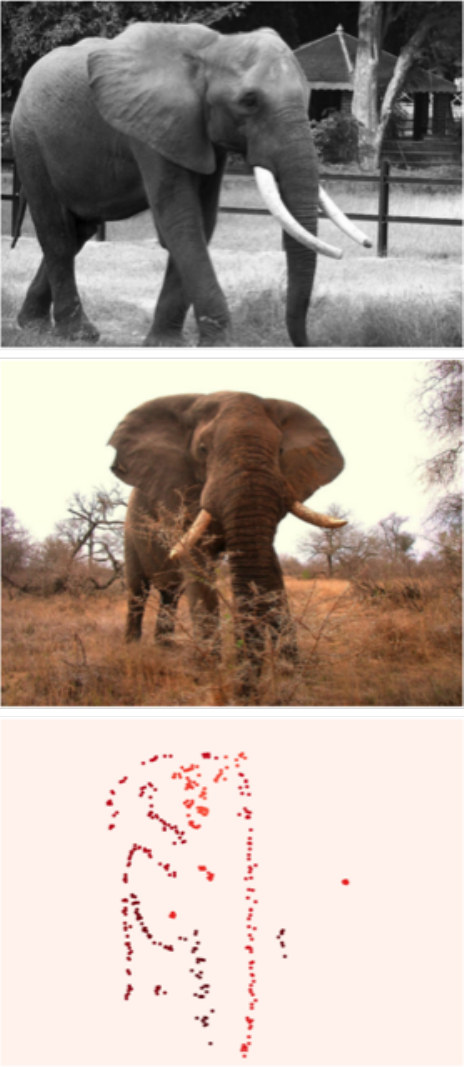}%
\includegraphics[width=0.2\textwidth]{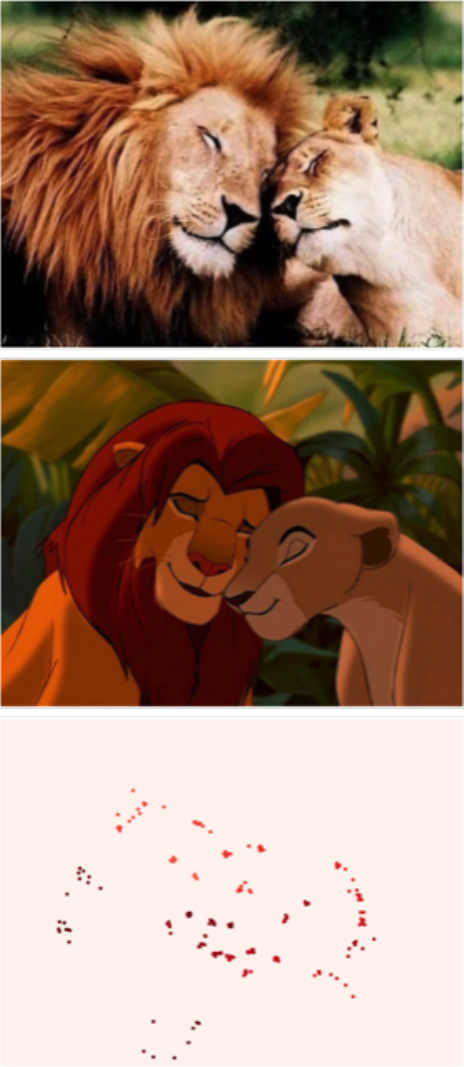}%
\includegraphics[width=0.2\textwidth]{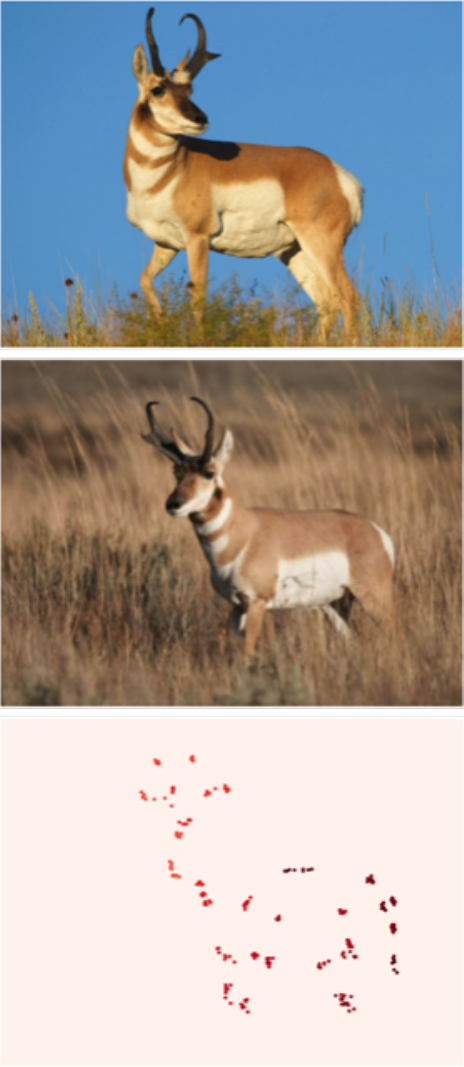}%
\includegraphics[width=0.2\textwidth]{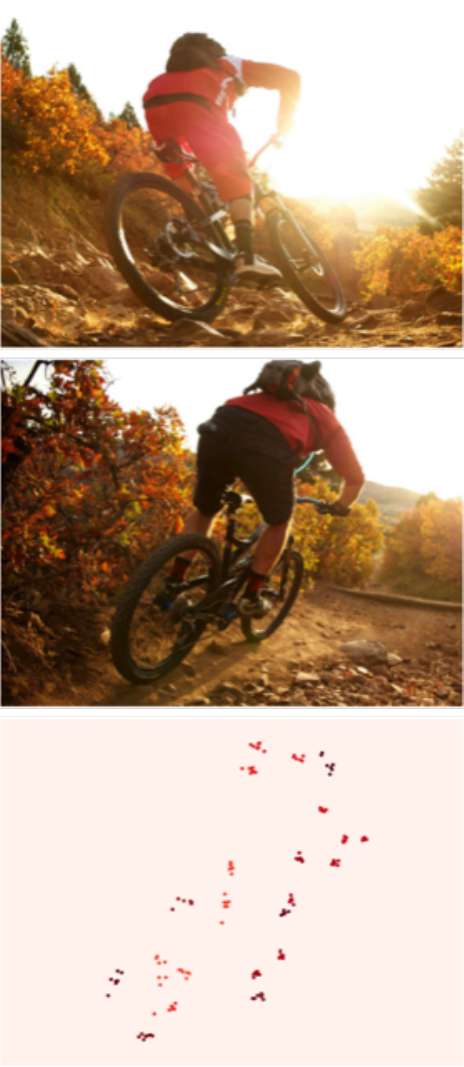}%
\includegraphics[width=0.2\textwidth]{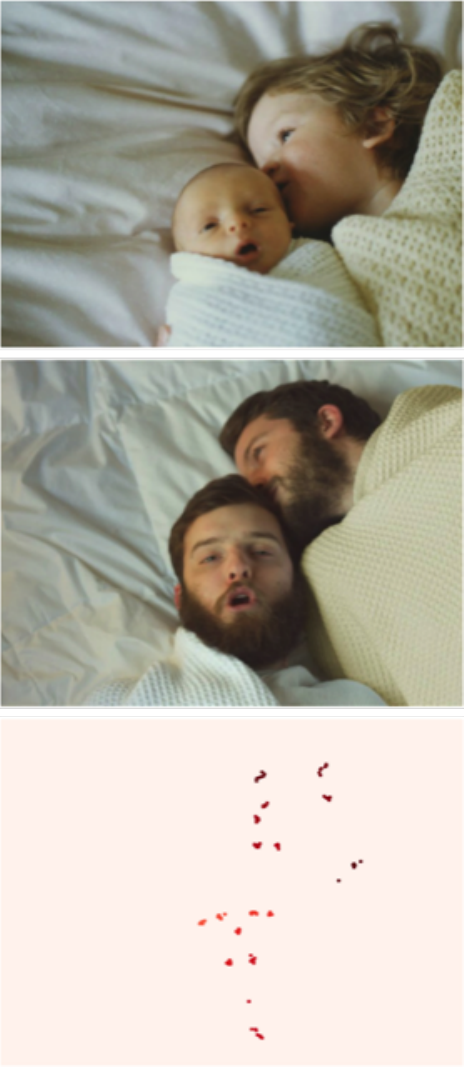}
\caption{A representative subset of the groundtruth dataset. From top to bottom: (a) the source images, (b) the target images, and (c) the distribution of points selected by the human annotators on the target images. The structure of the object is often clearly discernible from the annotations alone.
\label{fig:groundtruth}
}
\end{figure*}

In order to evaluate the efficacy of our method, we first wanted to understand how well human annotators perform at semantic labelling tasks. Since we are primarily interested in estimating correspondences for reconstruction-type objectives, we gathered 20 pairs of images from visual object categories which exhibit anatomical correspondence, including an assortment of animals, trucks, faces and people. Given a set of sparsely selected keypoints in the first image of each pair, 8 human annotators were tasked with labelling the corresponding points in the second image. A representative subset of the data is shown in~\fig{groundtruth}.

A similar experiment was performed in~\cite{liu_PAMI_2011}, however they focussed on correspondences across \emph{scenes}, which often have no clear correspondence, even for human annotators. In contrast, the agreement on our dataset is high, with a natural increase in uncertainty from corner features, to edges and textureless regions.

In recognizing that not all features are equally distinctive, we measure distance from estimated points $\x_i$ to the groundtruth using Mahalanobis distance,
\begin{align}
d_i(\x_i) = \sqrt{(\x_i - \mu_i)^T \S_i^{-1} (\x_i - \mu_i)}
\end{align}
where $\mu_i$ and $\S_i$ are the $2D$ mean and covariance of the groundtruth labellings across annotators. \cite{tompson_ARXIV_2015} motivate a similar procedure for human pose estimation. This metric has two advantages over Euclidean distance: (i) it takes into account spatial and directional uncertainty (\eg correspondences are afforded some slack along an edge, but not perpendicular to it), and (ii) it is resolution independent, since distance is measured in standard deviations.

Our dataset and metric therefore sets a higher standard for what is considered a good correspondence, both empirically and qualitatively (since readers can accurately discriminate good from poor results). All results presented in the following section are measured under this metric.

\subsection{Experiments}
In all of our experiments we resize the source (A) and target (B) image so
$\max(M,N) = 150$, preserving the aspect ratio, and extract densely sampled
SIFT features. 

The stationary distribution (mean and covariance) of SIFT features is estimated from $50,000$ randomly sampled images from ImageNet. Classifiers with spatial support $1 \times 1$, $3 \times 3$, $5 \times 5$, $7 \times 7$ and $9 \times 9$ were evaluated. The different sizes tradeoff speed, localization accuracy and generalization. We found $5 \times 5$ classifiers provided a good balance between these tradeoffs, and the results throughout our paper use this support.

While the LDA likelihoods are more computationally demanding to compute than $L_1$-norm likelihoods, the construction and application of the classifiers can be accelerated with BLAS. Estimating $10,000$ $5 \times 5$ classifiers and applying them in a sliding
window fashion to a $80 \times 125$ SIFT image (with 128 channels) takes
approximately 6 seconds.

\begin{figure*}
\includegraphics[width=0.33\textwidth]{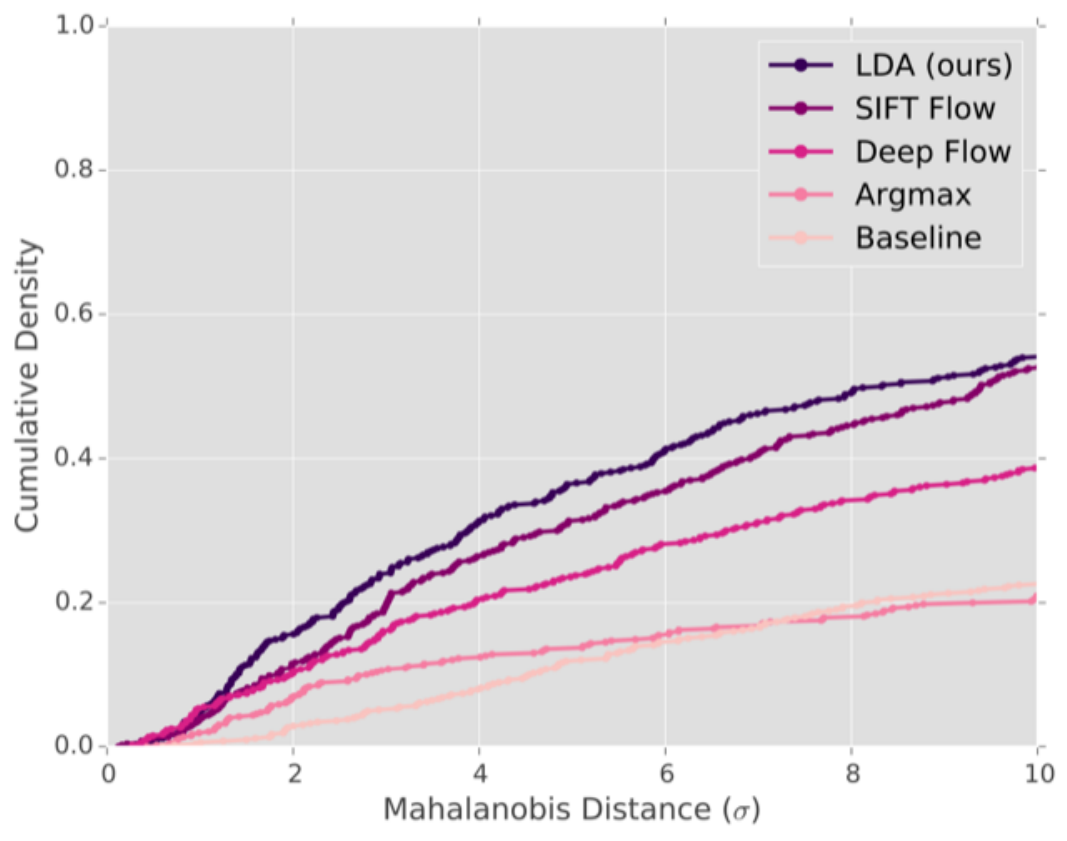}%
\includegraphics[width=0.33\textwidth]{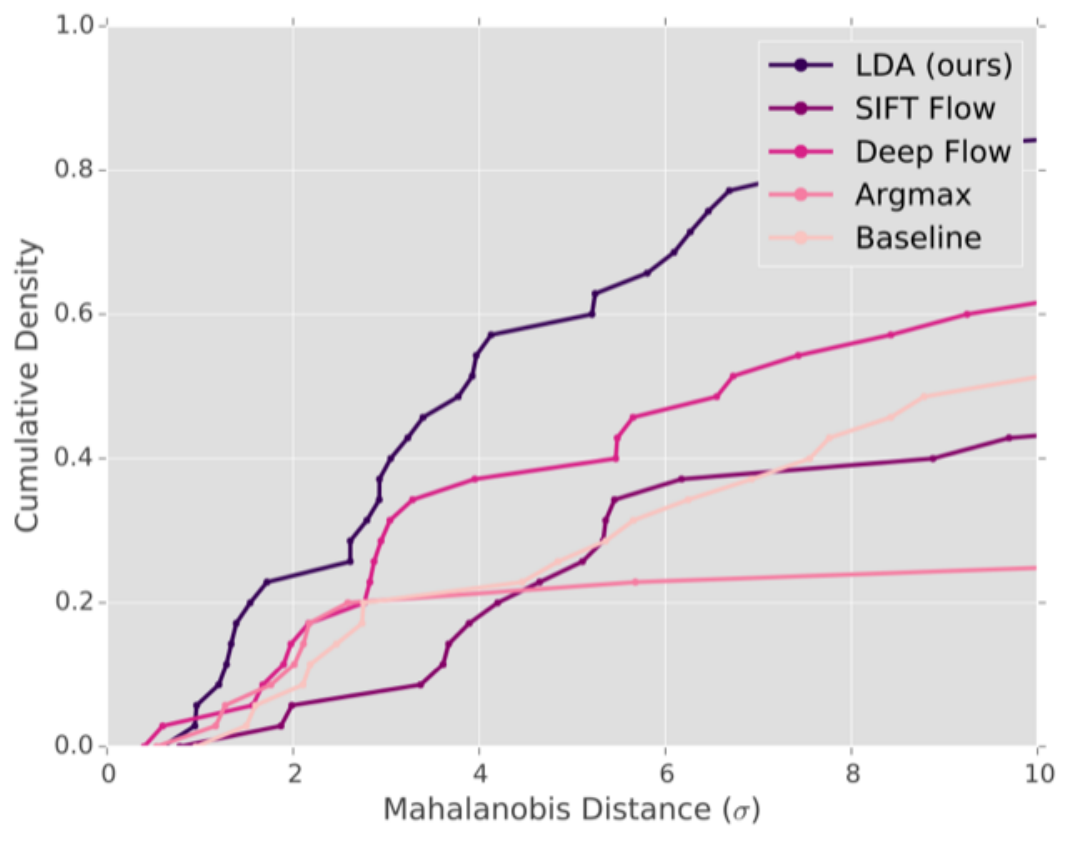}%
\includegraphics[width=0.33\textwidth]{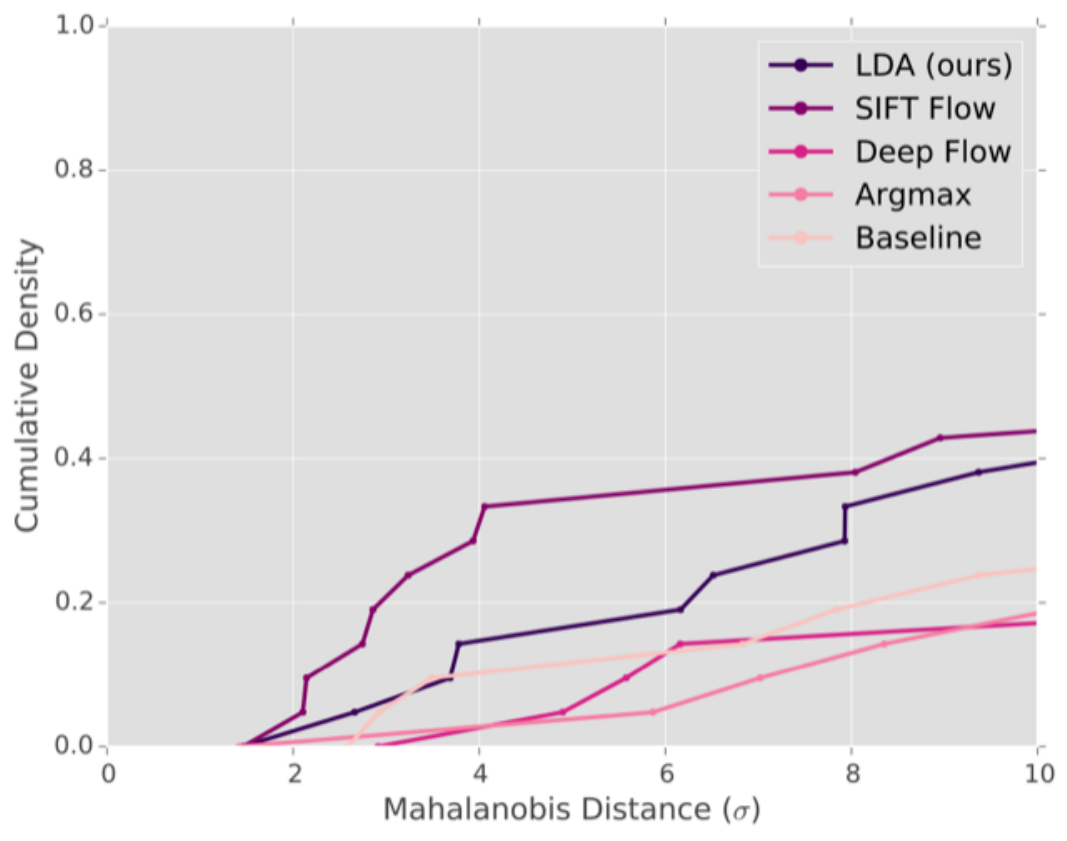}
\caption{Comparison of sparse keypoint localization for our method, SIFT Flow~\cite{liu_PAMI_2011} and Deep Flow~\cite{weinzaepfel_ICCV_2013}. The baseline measures the global alignment bias of the dataset (how well one would perform by simply assuming no flow). The argmax considers taking the single best match without regularization. The graphs measure the fraction of correspondences which fall within an increasing distance from groundtruth. $3$ standard deviations is inperceptible from human annotator accuracy. From left to right: (a) aggregate results across all images, (b) \textcolor[rgb]{0.157,0.647,0.827}{the truck pair} which our method localizes well, and \mbox{(c) \textcolor[rgb]{0.714,0.2,0.373}{the biking pair}} for which our method fails to produce any meaningful correspondences. 
\vspace{2mm}
\label{fig:results}
}
\end{figure*}

We apply our LDA-based correspondence method in the same graphical model framework as SIFT Flow. We use a coarse-to-fine scheme to handle inference over larger images, and grid searched the hyperparameters for both LDA and $L_1$ based unary functions. Results are shown in~\fig{results}.

We display the cumulative density for increasing number of standard deviations from groundtruth (\ie fraction of points falling within an increasing radius from groundtruth). As a baseline, we simply set \mbox{$\x_i = i$},\footnote{For images of different sizes, we set $\x_i = \W(i)$ where $\W$ is a function that maps the span of $\I_A$ to $\I_B$.} which acts as a proxy to the global alignment bias of the dataset (small flow assumption). In addition to SIFT Flow, we also compare our method to a leading optical flow method, Deep Flow~\cite{weinzaepfel_ICCV_2013}.

\begin{figure*}
\vspace{-1mm}
\begin{center}
\begin{minipage}{\textwidth}
\includegraphics[width=0.5\textwidth]{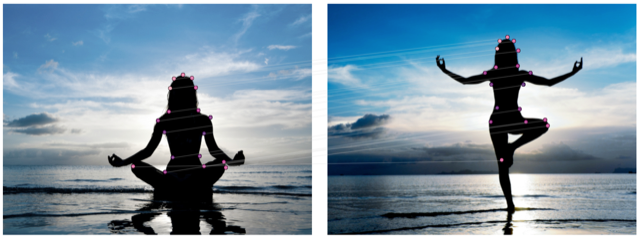}%
\includegraphics[width=0.5\textwidth]{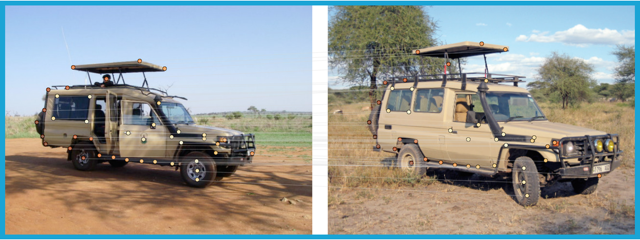}
\includegraphics[width=0.5\textwidth]{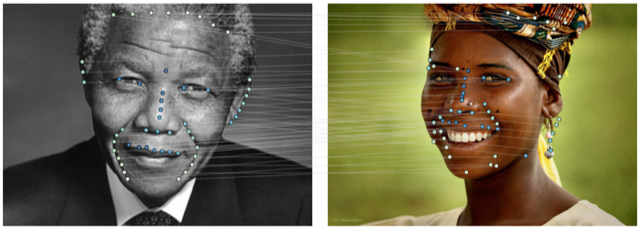}%
\includegraphics[width=0.5\textwidth]{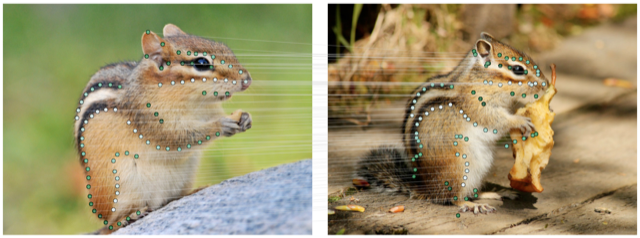}
\begin{minipage}{0.5\textwidth}
\includegraphics[width=\textwidth]{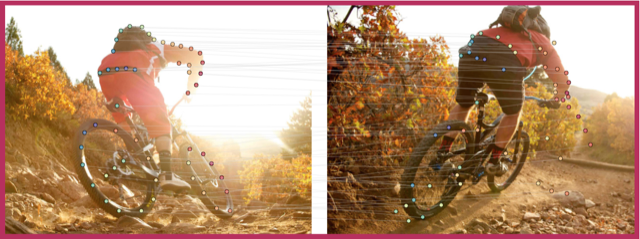}
\includegraphics[width=\textwidth]{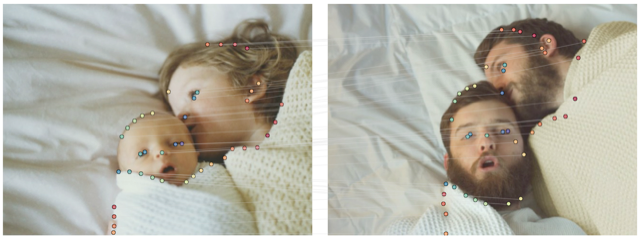}
\end{minipage}%
\begin{minipage}{0.5\textwidth}
\includegraphics[width=\textwidth]{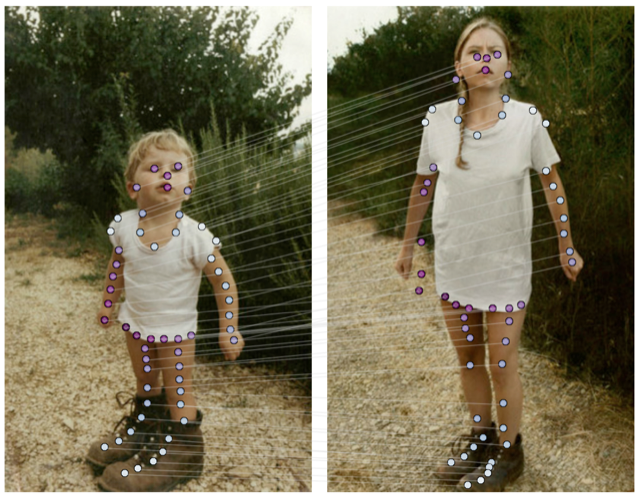}
\end{minipage}
\end{minipage}
\end{center}
\vspace{-3mm}
\caption{Example correspondences discovered by our method, across a broad range of image pairs from our dataset. The truck pair produces good localization of points (see \fig{results}b), whilst the biking pair shows a failure to produce anything meaningful (see \fig{results}c).
\label{fig:visualization}}
\end{figure*}

We truncate the CDF due to the long tails for all methods compared. This is an artefact of the non-global regularization schemes, which allow some points to be arbitrarily far from groundtruth without affecting others. Finally, in~\fig{visualization} we illustrate a number of exemplar correspondences to show the visual quality of matches produced by our method. 

%
%

%% file: 5-conclusion.tex
\section{Conclusion}
\label{sec:conclusion}

In this paper we motivated the application of dense semantic correspondence for a range of computer vision problems which currently rely on synthetic data or specialized imaging devices. In contrast to existing correspondence methods, which typically use similarity kernels, we proposed using exemplar classifiers for describing the likelihood of two points matching. We showed that LDA classifiers exhibit 3 desirable properties: (i) higher average precision than simple measures of image similarity such as the $L_1$ norm, (ii) significantly faster training than exemplar SVMs, and (iii) estimates of match confidence that are directly comparable across pixels.

We presented a small semantic correspondence dataset and metric in a bid to measure the performance of different methods in a quantifiable manner, and showed that under this metric our classifier-based approach offered improvements over the $L_1$ norm, within the same SIFT Flow optimization framework. The qualitative results illustrate our method's ability to estimate high-quality dense semantic correspondences.